# KERNEL INDUCED RANDOM SURVIVAL FORESTS

Fang Yang[a], Jiheng Wang[a] and Guangzhe Fan[a]
[a]Dept. of Statistics and Actuarial Science, Univ. of Waterloo, Waterloo, ON, Canada

## ABSTRACT

Kernel Induced Random Survival Forests (KIRSF) is a statistical learning algorithm which aims to improve prediction accuracy for survival data. As in Random Survival Forests (RSF), Cumulative Hazard Function is predicted for each individual in the test set. Prediction error is estimated using Harrell's concordance index (C index) [Harrell et al. (1982)]. The C-index can be interpreted as a misclassification probability and does not depend on a single fixed time for evaluation. The C-index also specifically accounts for censoring. By utilizing kernel functions, KIRSF achieves better results than RSF in many situations. In this report, we show how to incorporate kernel functions into RSF. We test the performance of KIRSF and compare our method to RSF. We find that the KIRSF's performance is better than RSF in many occasions.

## 1. INTRODUCTION

Survival data are often analyzed using methods that have restrictive assumptions such as Cox proportional hazard models (1972). Selecting variables or two-way and three-way interactions are very challenging and in practise, we must rely on experience and subjective knowledge to choose pertinent variables and their interactions. Usually, ad hoc approaches such as stepwise regressions are used to investigate if nonlinear effects or interactions are significant.

Tree models have been a useful alternative to Cox proportional hazard models for the exploration of survival data. Trees usually have fewer assumptions and can handle a large variety of data structures. Tree models adaptively partition the covariate space into regions and data into groups. For a categorical covariate, the spit has the form of "$X_j \leq c$ or $X_j > c$". Some measure of separation in the response distribution such as likelihood ratio test statistic between the two groups is calculated. The covariate and the split point that best separate the two groups are chosen and the same procedure is applied to the subgroups until many disjoint groups has been formed. Several tree based models has been proposed to analyze right censored survival data. Segal [1] uses log-rank statistics as split functions to build survival trees. Davis and Anderson [2] use an exponential log-likelihood split criterion to build an exponential survival tree model. Leblanc and Crowley [3] proposed a relative risk tree model. However, tree models have the disadvantage of instability due to their greedy search for best predictors, for best splits and search multiple times.

Tree ensemble methods such as random survival forests (RSF) developed by Ishwaran, et al [4] can solve the problem of instability of a single survival tree model. RSF consists of many survival trees. The construction of each tree involves two randomizations. One is to randomly draw a bootstrap sample from data. The other is to randomly draw a subset of predictors. The final decision is based on the average of prediction results from all the individual trees. According to Ishwaran, et al [4], prediction error is estimated using Harrell's concordance index (C index) [Harrell et al. (1982)] because C index can be interpreted as a misclassification probability.

In this paper, we take advantage of kernel functions which allow linear learning classifiers to perform non-linear learning. The aim is to improve prediction accuracy of RSF. We observed that our method improves RSF greatly in terms of prediction accuracy.

## 2. COX MODEL

One of the important problems in survival analysis is to predict the distribution of the time to some event given a set of covariates. Cox proportional hazards model is also called Cox model or sometimes semi-parametric regression model. In Cox model, it assumes that the hazard functions of two individuals are proportional for all values of time $t$, i.e.

$$\frac{\lambda_i(t)}{\lambda_j(t)} = \frac{\lambda_0(t)e^{\beta' z_i}}{\lambda_0(t)e^{\beta' z_j}} = e^{\beta'(z_i - z_j)} \text{ for all } t>0 \tag{1}$$

where $\lambda_i(t)$ is the hazard function for individual $i$ at time $t$; $\lambda_0(t)$ is the unspecified baseline hazard; $z_i$ is the px1 vector of covariates for individual $i$; $\beta$ is px1 vector of unknown parameter.

According to Cox (1972, 1975), inference about $\beta$ can be done without worrying about $\lambda_0(t)$, which means $\beta$ can be estimated by maximizing the partial likelihood function instead of the true full likelihood function. The partial likelihood function has the following form:

$$L(\beta) = \prod_{i=1}^{n} \left( \frac{e^{\beta' z_i}}{\sum_{j=1}^{n} y_j(t_i) e^{\beta' z_j}} \right)^{\delta_i} \tag{2}$$

Here we assume that we have a random sample of $n$ possibly right censored survival times $(t_1, \delta_1)$, $(t_2, \delta_2)$, ⋯, $(t_n, \delta_n)$. $\delta_i$ =1 if individual $i$ is not censored; 0 otherwise for all $i$ =1,2, ⋯, $n$. $y_j(t) = I\,(t_j \geq t)$ , which is equal to 1 if individual $j$ is still under study at time $t$, 0 otherwise. The partial likelihood formulas implies that censored individuals does not contribute to the partial likelihood at the time they are censored, but do contribute to the partial likelihood at other time points through the risk set (Risk set includes all individuals under study at a certain time $t$). Relative risk or hazard ratio can be obtained after parameter vector $\beta$ are estimated. The relative risk or hazard ratio is the ratio of the hazard functions that corresponds to a change of one unit for a given covariate while the other covariates remain the same. The general form is:

$$\mathrm{RR}_{\mathrm{k}} = \exp\,(\beta_k\,),\ \mathrm{k} = 1, 2, \cdots, p \tag{3}$$

When number of covariates is large, it is a big challenge to decide which interactions should be considered to be included in the model. However, tree models can handle interactions automatically. In next section, we discuss about survival tree models.

## 3. RELATIVE RISK SURVIVAL TREE

Tree models have been a useful alternative to Cox proportional hazard models for the exploration of survival data. Trees usually have fewer assumptions and can handle a large variety of data structures. Tree models adaptively partition the covariate space into regions and data into groups. For a categorical covariate, the spit has the form of "$X_j \leq c$ or $X_j > c$". Some measure of separation in the response distribution such as likelihood ratio test statistic between the two groups is calculated. The covariate and the split point that best separate the two groups are chosen and the same procedure is applied to the subgroups until many disjoint groups has been formed.

Many methods have been proposed to construct a survival tree. We will focus on relative risk survival tree in this section. Same as Cox proportional hazard model, relative risk survival trees assume that hazard rate could change over time, but is proportional to the baseline hazard. The hazard rate for individual i at time t has the following form:

$$\lambda_i(t) = \lambda_0(t)\theta_i \tag{4}$$

where $\theta_i \geq 0$ and $\lambda_0(t)$ is the baseline hazard at time $t$ for all individuals.

The key difference between relative risk survival tree and Cox model is that Cox model assumes an exponential form for the relative risk part, which is $e^{\beta' z_i}$. On contrary, relative risk survival tree does not specify any form for $\theta_i$. $\theta_i$ is just an unknown parameter. Hence, we say that relative risk survival tree is a non-parametric model, but Cox model is a semi-parametric model. The aim of the relative risk tree model is to estimate relative risk $\theta_i$ for individual $i$. An example of a binary tree structure with three terminal nodes is shown in Figure 1.

For a relative risk survival tree, in each node, the splitting covariate and splitting value are determined by maximizing deviance deduction. The deviance in node $k$ can be obtained by the following procedures.

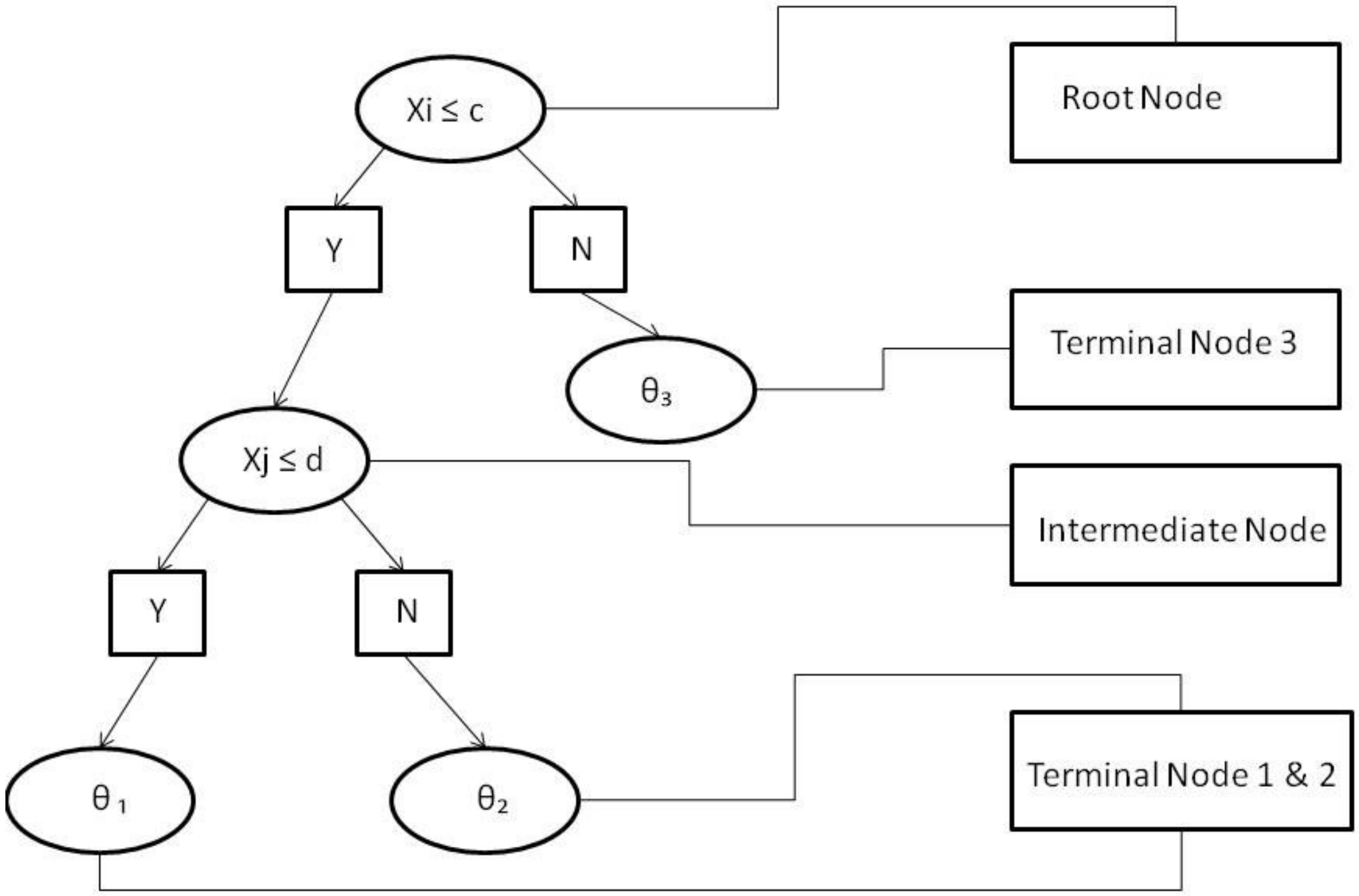

**Figure 1. An example of a binary tree structure with three terminal nodes**

We assume that $T$ is the time under observation and $\delta$ is an indicator of failure. $X = (X_1, X_2, \cdots, X_m)$ are a vector of $M$ covariates. Suppose $U$ is the true survival time and $V$ is the true censoring time, $T = \min(U, V)$. Also, assume that the $U$ and $V$ are independent given $X$. The full likelihood of the training sample for a tree $T$ can be expressed as

$$L = \prod_{h \in \hat{T}} \prod_{i \in S_h} \left(\lambda_0(t_i)\theta_h\right)^{\delta_i} e^{-\Lambda_0(t_i)\theta_h} \tag{5}$$

Where $\Lambda_0(t)$ is the baseline cumulative hazard function; $\lambda_0(t)$ is hazard function; $\theta_h \geq 0$, $\check{T}$ is the set of terminal nodes; $S_h$ is the set of observation labels in terminal node $h$. Then the likelihood estimate for $\theta_h$ is

$$\hat{\theta}_h = \frac{\sum_{i \in S_h} \delta_i}{\sum_{i \in S_h} \Lambda_0(t_i)} \tag{6}$$

In practice, $\Lambda_0(t)$ is unknown, but can be estimated numerically. An extensive discussion can be found at [3]. Then, the deviance in node $k$ is defined as:

$$D(k) = \sum_{i \in S_k} 2\left\{ \delta_i \log\left( \frac{\delta_i}{\hat{\Lambda}_0(t_i)\hat{\theta}_k} \right) - \left(\delta_i - \hat{\Lambda}_0(t_i)\right)\hat{\theta}_k \right\} \tag{7}$$

where $s_k$ is the set of all observations falling in node $k$, and are the maximum likelihood estimations of $\theta_k$ and the baseline cumulative hazard function.

For individuals that fall in the same terminal node $k$ ($k$ = 1, 2, 3 in this case), we say that they have the same relative risk $\theta_k$. Also, we can obtain the relative risk between two different groups who fall into two different terminal nodes $i$ and $j$ by $\frac{\theta_i}{\theta_j}$.

Although tree model can handle interactions automatically, it is a very unstable method due to its greedy search for best covariate and best split value. Adding one noise point in the data set can change a tree model structure significantly. As well known, tree ensemble methods are very stable and can achieve high prediction accuracy.

# 4. RANDOM SURVIVAL FORESTS

Random Survival Forests (RSF) is an ensemble tree method for right-censored survival data. RSF consists of many survival trees. The final result is based on the average of the results from all of the individual survival trees. Each survival tree is a binary tree which is grown using recursive splitting rules. For example, a log-rank splitting rule is to search all possible covariates and split values $c$ and then find a covariate (should be coded as discrete) $x^*$ and a value $c^*$ that can maximize the log-rank test statistic at each node. In other words, the best split in each node is found by maximizing survival difference. Eventually, as the tree grows, the terminal node is populated by individuals with similar survival. Let $(T_{1,h}, \delta_{1,h})$, $(T_{2,h}, \delta_{2,h})$, $\cdots$, $(T_{n(h),h}, \delta_{n(h),h})$ be the survival times and censoring status in a given terminal node $h$. An individual $i$ is said to be censored if $\delta_{i,h}$=0 and have died (or experienced an event) if $\delta_{i,h}$=1. Let $t_{1,h} < t_{2,h} < \cdots < t_{N(h),h}$ be the $N(h)$ distinct event times, $d_{i,h}$ be the number of death and $r_{i,h}$ be the number of individuals at risk at time $t_{i,h}$.

Then the Nelson-Aalen estimator for cumulative hazard function (CHF) is defined as

$$\hat{H}_h(t) = \sum_{t_{i,h} \le t} \frac{d_{i,h}}{r_{i,h}} \tag{8}$$

All individuals within the terminal node $h$ have the same CHF.

Suppose individual $i$ has a d-dimensional covariate $x_i$. Then the Nelson-Aalen estimator for CHF for individual $i$ is $H(t|x_i) = \hat{H}_h(t)$ when $i$ falls into terminal node $h$. According to Ishwaran, et al. [2008], the bootstrap ensemble CHF for $i$ is

$$H_e(t|x_i) = \frac{1}{B}\sum_{b=1}^{B} H_b(t|x_i) \tag{9}$$

Where $H_b(t|x_i)$ denotes the CHF (4.1) for the tree grown from the $b$-th bootstrap sample and $H_e(t|x_i)$ denotes the bootstrap ensemble CHF. Following Ishwaran, et al [4], *let* $(T_1, \delta_1), (T_2, \delta_2), \cdots, (T_n, \delta_n)$ denote survival times and censoring status for the non-bootstrapped data, the ensemble mortality for i is defined as

$$\hat{M}_{e,i} = \sum_{j=1}^{n} H_e(T_j|x_i) \tag{10}$$

Prediction error can be estimated by using Harrell's concordance index (C index) [Harrell et al. (1982)] because C index can be interpreted as a misclassification probability [4]. The C-index (concordance index) is related to the area under the receiver operating characteristic (ROC) curve [Heagerty and Zheng (2005)]. It estimates the probability that, in a randomly selected pair of cases, the case that fails first had a worst predicted outcome. The interpretation of the C-index as a misclassification probability is attractive, and is one reason they use it for prediction error. Another attractive feature is that, unlike other measures of survival performance, the C-index does not depend on a single fixed time for evaluation. The C-index also specifically accounts for censoring.

The C-index is calculated using the following steps:

1. Form all possible pairs of cases over the data.

2. Omit those pairs whose shorter survival time is censored. Omit pairs $i$ and $j$ if $T_i = T_j$ unless at least one is a death. Let Permissible denote the total number of permissible pairs.

3. For each permissible pair where $T_i \neq T_j$, count 1 if the shorter survival time has worse predicted outcome; count 0.5 if predicted outcomes are tied. For each permissible pair, where $T_i = T_j$ and both are deaths, count 1 if predicted outcomes are tied; otherwise, count 0.5. For each permissible pair where $T_i = T_j$, but not both are deaths, count 1 if the death has worse predicted outcome; otherwise, count 0.5. Let Concordance denote the sum over all permissible pairs.

4. The C-index, C, is defined by C = Concordance/Permissible.

Calculating C requires a predicted outcome. Ishwaran, et al [4] use the OOB ensemble CHF to define a predicted outcome. Because this value is derived from OOB data, it can be used to obtain an OOB estimate for C, and, consequently, an OOB error rate.

Let $t_1^o, \cdots, t_m^o$ denote pre-chosen unique time points (we use the unique event times, $t_1, \cdots, t_N$ ). To rank two cases *i* and *j*, we say *i* has a worse predicted outcome than *j* if

$$\sum_{l=1}^{m} H_e^{**}\left(t_l^o \middle| x_i\right) \succ \sum_{l=1}^{m} H_e^{**}\left(t_l^o \middle| x_j\right) \tag{11}$$

where $H_e^{**}\left(t \middle| x_i\right) = \dfrac{\sum_{b=1}^{B} I_{i,b} H_b^*(t|x_i)}{\sum_{b=1}^{B} I_{i,b}}$ and $H_b^*\left(t \middle| x\right)$denote CHF (8) for a tree grown from *b*-th bootstrap sample.

Using this rule, compute C as outlined above. Denote the OOB estimate by C**. The OOB prediction error, PE**, is defined as 1 – C**. Note that 0 ≤ PE** ≤ 1 and that a value PE** = 0.5 indicates prediction no better than random guessing.

In this paper, we aim to decrease prediction error by introducing kernel functions to Random Survival Forests. In next section, we show how to incorporate kernel functions into Random Survival Forests.

# 5. KERNEL INDUCED RANDOM SURVIVAL FORESTS

## 5.1 Kernel Functions

In the last decade there has been a surge in the number of statistical algorithms that have been shown to benefit from kernels. Kernel Principal Components Analysis (kPCA) [9] and the support vector machine (SVM) [10] are two typical statistical methods which use kernels to study high dimensional data. Although tree models are not linear classifier as PCA and SVM, we still believe it will benefit from kernel methods.

*Kernel* functions can map data to higher dimension and then take inner product without explicitly writing out the new feature space. For example, we assume there are two covariates $X_1$ and $X_2$. For individuals *i* and *j,* $\vec{x}_i = \left(X_{i1}, X_{i2}\right)$ and $\vec{x}_j = \left(X_{j1}, X_{j2}\right)$. Let's define the kernel function as

$$K\left(\vec{x}_i, \vec{x}_j\right) = \left\langle \phi\left(\vec{x}_i\right), \phi\left(\vec{x}_j\right) \right\rangle = \left\langle \vec{x}_i, \vec{x}_j \right\rangle^2 \tag{12}$$

where $\left\langle \vec{x}_i, \vec{x}_j \right\rangle^2$ denotes the inner product of $\vec{x}_i$ and $\vec{x}_j$ and hence

$$\begin{aligned}\langle \vec{x}_i, \vec{x}_j \rangle^2 &= (X_{i1}X_{j1} + X_{i2}X_{j2})^2 \\ &= (X_{i1}^2 X_{j1}^2 + 2X_{i1}X_{j1}X_{i2}X_{j2} + X_{i2}^2 X_{j2}^2) \\ &= (X_{i1}^2, \sqrt{2}X_{i1}X_{i2}, X_{i2}^2)(X_{j1}^2, \sqrt{2}X_{j1}X_{j2}, X_{j2}^2) \\ &= \langle \phi(\vec{x}_i), \phi(\vec{x}_j) \rangle\end{aligned}$$

Therefore, the kernel function maps feature space $(X_1, X_2)$ to feature space $(Z_1, Z_2, Z_3) = (X_1^2, \sqrt{2}X_1X_2, X_2^2)$, which is mapping $R^2 \rightarrow R^3$. A linear model on the new feature space is thus equivalent to a non-linear model on the original feature space. The great advantage of kernel methods is that we don't have to find the new feature space explicitly because they could be even infinite. The kernel functions we use in this paper are polynomial kernel and Gaussian kernel. A polynomial kernel with degree $d \geq 2$ is

$$K(\vec{x}_i, \vec{x}_j) = (\langle \vec{x}_i, \vec{x}_j \rangle + c)^d \tag{13}$$

where $c$ is a tuning parameter. A Gaussian kernel is defined as

$$K(\vec{x}_i, \vec{x}_j) = e^{\left\{\frac{-\|\vec{x}_i - \vec{x}_j\|^2}{2\sigma^2}\right\}} = e^{\left\{\frac{-(\langle \vec{x}_i, \vec{x}_i \rangle + \langle \vec{x}_j, \vec{x}_j \rangle - 2\langle \vec{x}_i, \vec{x}_j \rangle)}{2\sigma^2}\right\}} \tag{14}$$

where $\sigma^2$ is a tuning parameter. The Gaussian kernel is also referred to as a radial basis function (RBF) kernel. Kernel methods are suitable for data sets which have non-linear patterns. By performing linear analysis in higher dimension, it is equivalent to perform non-linear analysis in lower dimension.

## 5.2 Kernel-Induced Random Survival Forests

Suppose we have $n$ observed individuals. Each individual is characterized by $d$ covariates. We know that each individual $i$ could be related to other individuals via a *kernel* function $K(\vec{x}_i, \cdot)$. The value of this kernel function is uniquely determined by the given individual vector $\vec{x}_i$ and another individual vector $\vec{z}$. We can view such a *kernel* function as a new covariate, defined as $K_i(\vec{z}) = K(\vec{x}_i, \vec{z})$. The new covariate is observation-based and *kernel*-induced, i.e., if we consider all $n$ observations in the learning set, then we will have $n$ such *kernel*-induced covariates, $K_i(\cdot)$, $i$=1, 2, 3,$\cdots$, $n$. These new covariates or features are different from the inexplicitly new covariates or features that kernel functions imply. Since the new features are kernel induced, we call this method as kernel induced random survival forests (KIRSF). KIRSF applies the same algorithms as random survival forests. The key difference is that we use $n$ by $n$ dimensional new kernelized data instead of the $n$ by $d$ dimensional original data. In next section, we compare the performance of KIRSF to RSF.

# 6. EXAMPLES AND RESULTS

## 6.1 A simulated example

Suppose that the survival time of the population of interest follows exponential distributions with the rate parameters of $\lambda_1$ =0.1 and $\lambda_2$=0.5. There are 20 covariates which are generated by using Ringnorm in the "mlbench" library in R. The survival times are censored by a uniform distribution [5, 10]. We simulated 1,000 observations and randomly chose 100 as training set to build models. Then the trained models are used to obtain the predicted values for the cumulative hazard functions for the unused 900 observations. The first six observations are shown below:

| | c | t | X3 | X4 | X5 | X6 | X7 | X8 | X9 | X10 | X11 | X12 | X13 | X14 | X15 | X16 | X17 | X18 | X19 | X20 | X21 | X22 |
|---|---|---|---|---|---|---|---|---|---|---|---|---|---|---|---|---|---|---|---|---|---|---|
| 1 | | 2.398 | 2.279 | -2.130 | 0.848 | 2.040 | 0.054 | 0.760 | -2.277 | 0.554 | 1.081 | -2.893 | -3.692 | -1.309 | 1.031 | -1.521 | -1.396 | 0.722 | 1.060 | -0.975 | 3.058 | -1.701 |
| 2 | 0 | 5.729 | -1.392 | -2.620 | 1.004 | -2.768 | -0.894 | 1.307 | -0.768 | 1.921 | 4.523 | -0.204 | 3.240 | 1.903 | 2.799 | -1.832 | 1.074 | -2.081 | -1.395 | 1.110 | -0.232 | 0.790 |
| 3 | 1 | 0.172 | -2.127 | 0.285 | -0.635 | 1.551 | -2.244 | 1.235 | -1.185 | 0.833 | 0.036 | -1.950 | 0.077 | 1.958 | 3.016 | -0.001 | 0.575 | -0.042 | 0.780 | 1.818 | 0.043 | -4.690 |
| 4 | 0 | 7.395 | -0.442 | 0.539 | -3.406 | -1.868 | 2.101 | 0.078 | 6.724 | -2.093 | -1.137 | 0.262 | -0.199 | -0.902 | -1.438 | -2.519 | 0.011 | 1.816 | -1.798 | -3.022 | -2.842 | -1.953 |
| 5 | 0 | 7.901 | -5.119 | 0.157 | 0.922 | 0.095 | -1.088 | 0.003 | -0.003 | 1.914 | -0.070 | -4.578 | -3.681 | 1.421 | -0.868 | 0.459 | -0.472 | 1.640 | -0.943 | -2.096 | -1.928 | -2.699 |
| 6 | 1 | 1.935 | 2.128 | 2.689 | 1.575 | -0.866 | -1.677 | 0.188 | -1.200 | -0.457 | -2.139 | 0.759 | -0.661 | 0.802 | -0.790 | -0.404 | 0.906 | 1.900 | 2.783 | 2.337 | 0.308 | -0.201 |

Gaussian Kernel is a good choice for KISRF here. The experiment is done with 50 realizations using different random seeds. The results are summarized in Table 1 and the box plot of the error rates with 50 realizations is shown in Figure 2.

**Table 1. Summarized results of Simulation 3.1. The average performances (standard deviation) over 50 random realizations are shown.**

| Method | Mean error rate % | Sample Standard deviation % |
|---|---|---|
| KIRSF | 33.23 | 0.5158 |
| RSF | 41.8 | 3.01 |

According to the results of two sample *t*-test, *t* value is 19.82, degree of freedom is 98 and p-value is smaller than 0.0001. We conclude that KIRSF improves RSF significantly and performs better not only in terms of accuracy, but also in precision. The survival functions for each individual in the training set using RSF and KIRSF are shown in Figure 3 and Figure 4, respectively. The overall ensemble survival, Nelson-Aalen estimator and the real survival functions are also shown in these two figures.

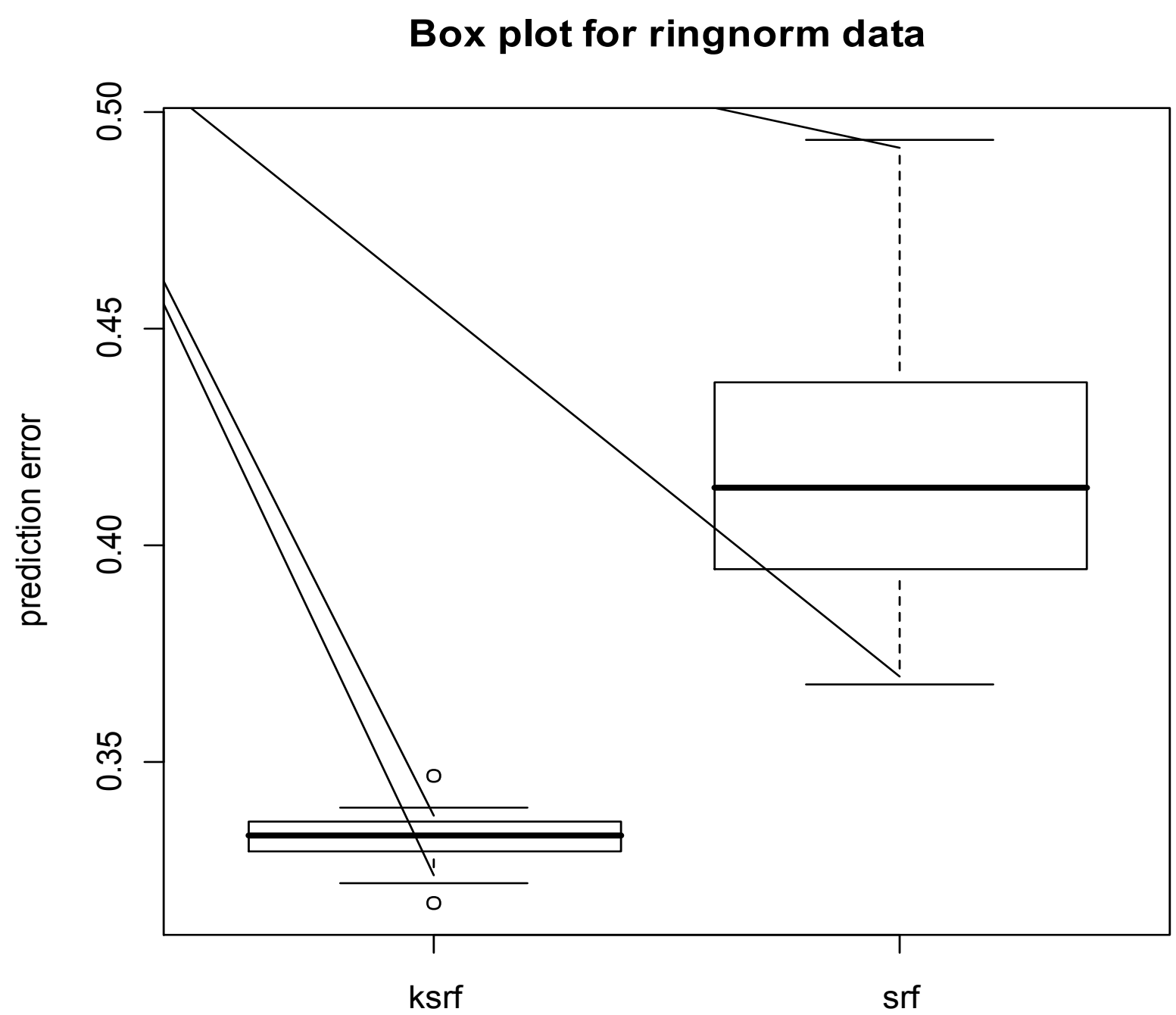

**Figure 2. The error rates with 50 realizations**

Figure 3 and Figure 4 show that both random survival forests overall ensemble survival and kernel introduced random survival forests overall ensemble survival are very close to the Nelson-Aalen estimator. However, it is clear that there is a gap among the kernel induced random survival function for each individual in the training set, which indicates that there are two possible different survival groups. Therefore, we conclude that kernel induced random survival functions emulated the nature more accurately because the true population which we used to generate the simulated data have two different survival distributions.

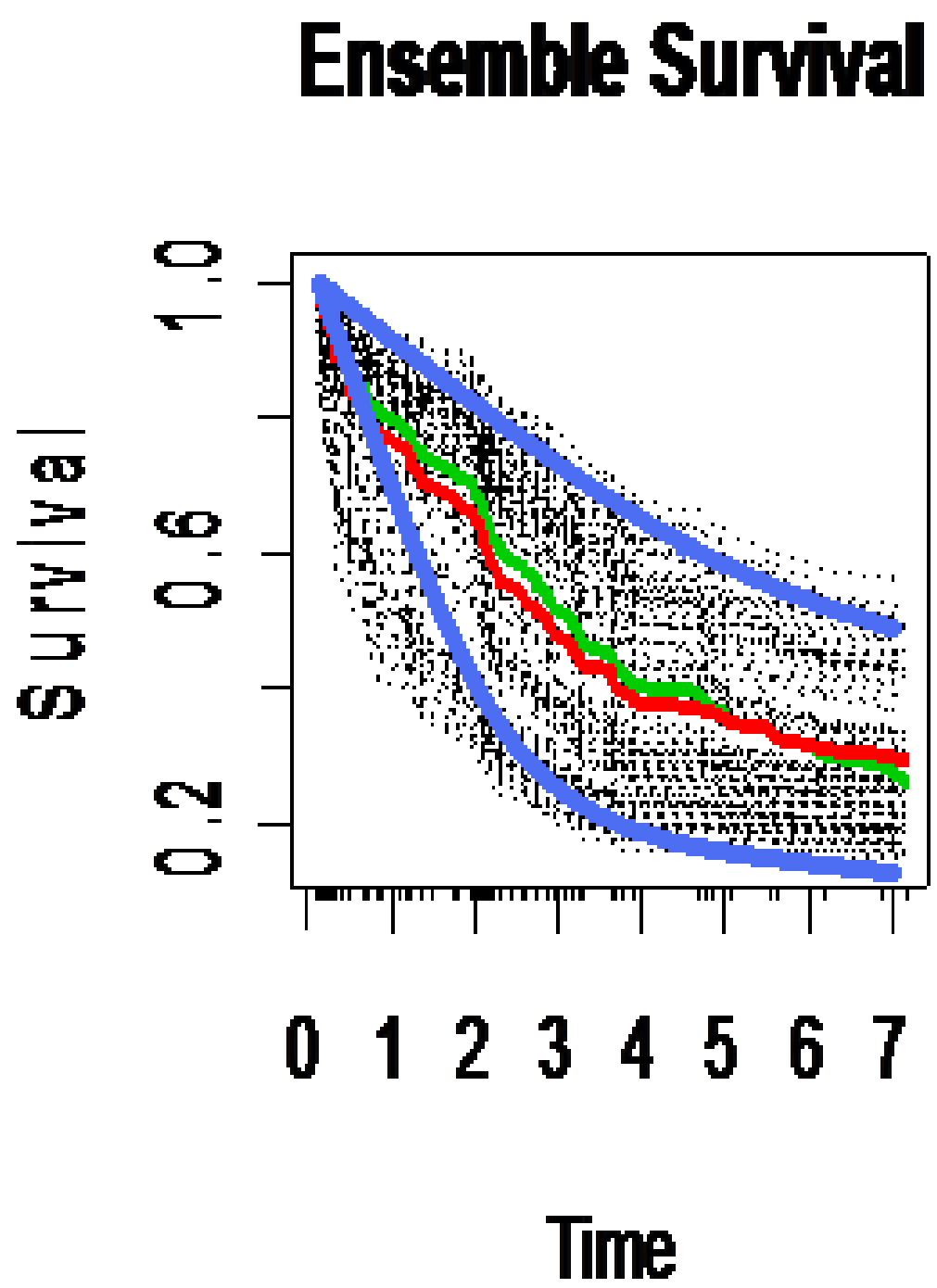

**Figure 3. Survival functions: Red line is RSF overall ensemble survival. Green line is Nelson-Aalen estimator. Black lines are RSF survival functions for each individual in the training set. Blue lines are the real survival functions.**

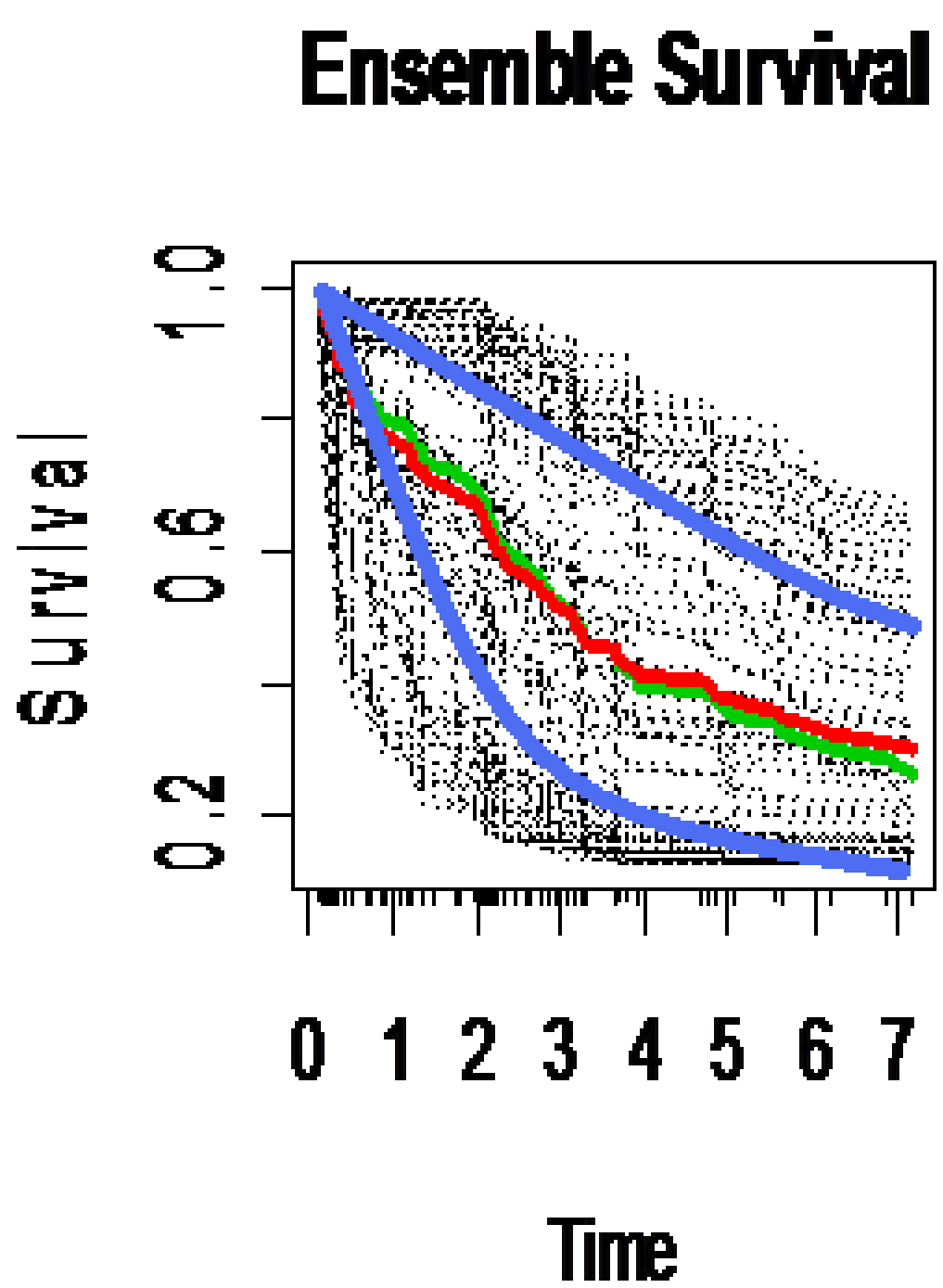

**Figure 4. Survival functions: Red line is KIRSF overall ensemble survival. Green line is Nelson-Aalen estimator. Black lines are KIRSF survival functions for each individual in the training set. Blue lines are the real survival functions.**

## 6.2 BMT data

This data set is from Medical College of Wisconsin in bone marrow transplant treatment for acute leukemia conducted in four hospitals from 1984 to 1989. The full description of the data can be found on the website: http://www.mcw.edu/biostatistics/Faculty/Faculty/JohnPKleinPhD/SurvivalAnalysisBook/DataSetsBothEditions.htm. There are 137 patients. The data for the first six patients are shown in Table 2.

**Table 2. The data for the first six patients**

| ID | c | t | ta | a | tc | c | tp | p | Z1 | Z2 | Z3 | Z4 | Z5 | Z6 | Z7 | Z8 | Z9 | Z10 | Group |
|---|---|---|---|---|---|---|---|---|---|---|---|---|---|---|---|---|---|---|---|
| 1 | 0 | 2081 | 67 | 1 | 121 | 1 | 13 | 1 | 26 | 33 | 1 | 0 | 1 | 1 | 98 | 0 | 1 | 0 | 1 |
| 2 | 0 | 1602 | 1602 | 0 | 139 | 1 | 18 | 1 | 21 | 37 | 1 | 1 | 0 | 0 | 1720 | 0 | 1 | 0 | 1 |
| 3 | 0 | 1496 | 1496 | 0 | 307 | 1 | 12 | 1 | 26 | 35 | 1 | 1 | 1 | 0 | 127 | 0 | 1 | 0 | 1 |
| 4 | 0 | 1462 | 70 | 1 | 95 | 1 | 13 | 1 | 17 | 21 | 0 | 1 | 0 | 0 | 168 | 0 | 1 | 0 | 1 |
| 5 | 0 | 1433 | 1433 | 0 | 236 | 1 | 12 | 1 | 32 | 36 | 1 | 1 | 1 | 1 | 93 | 0 | 1 | 0 | 1 |
| 6 | 0 | 1377 | 1377 | 0 | 123 | 1 | 12 | 1 | 22 | 31 | 1 | 1 | 1 | 1 | 2187 | 0 | 1 | 0 | 1 |

We are interested in the treatment related mortality or death in remission without their platelets returning to normal levels. We randomly split the data into training and test set with 90% and 10% of the observations, respectively. Linear kernel is used and the experiment is done with 100 realizations. The results are summarized in Table 3 and the box plot of the error rates with 100 realizations is shown in Figure 5.

**Table 3. Summarized results of example 6.2. The average performances and standard deviation over 100 random realizations are shown.**

| Method | Mean error rate % | Sample Standard deviation % |
|---|---|---|
| KIRSF | 12.20 | 7.02 |
| RSF | 18.11 | 8.13 |

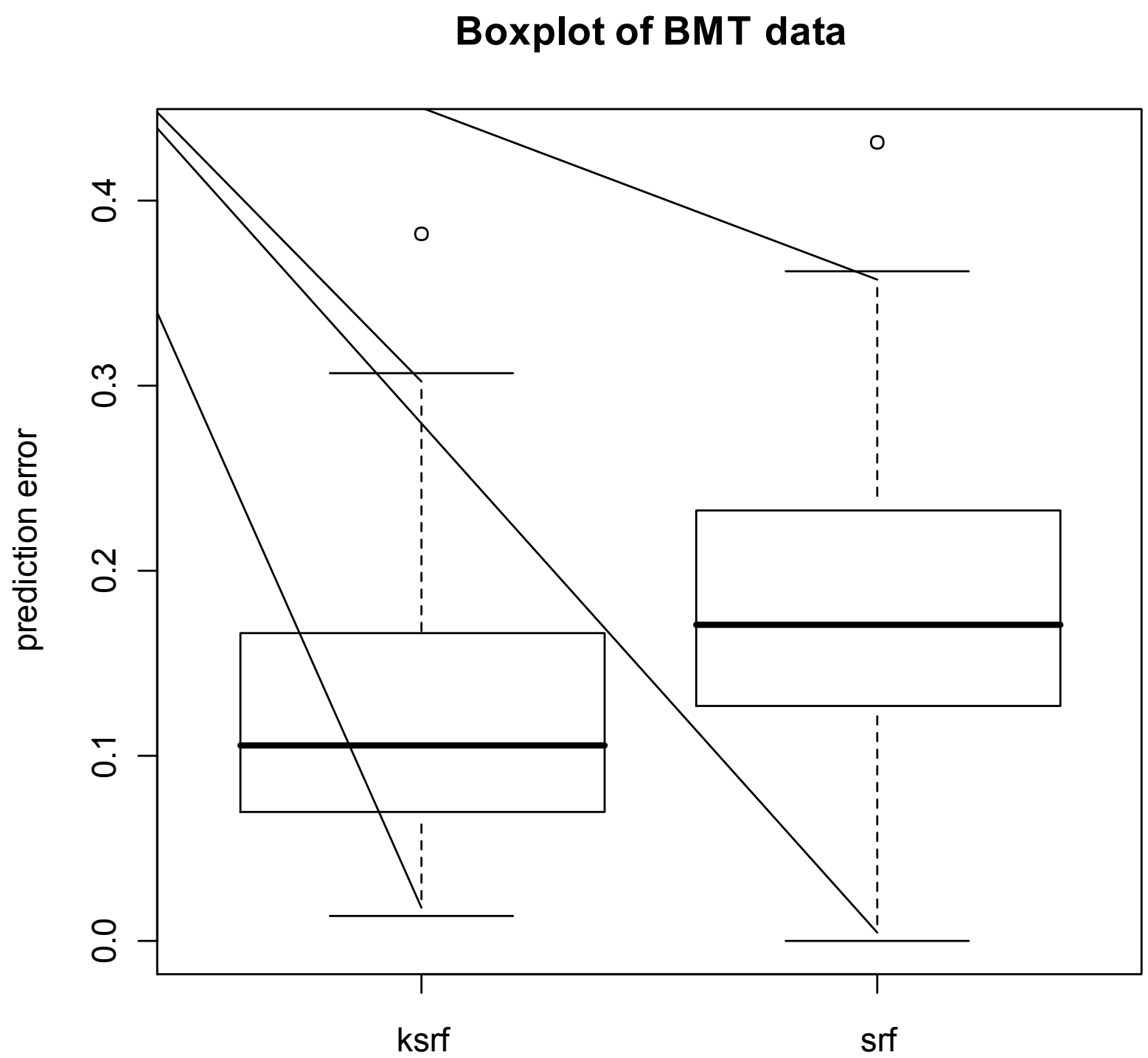

**Figure 5. The error rates with 100 realizations**

According to two sample t-test results, the *t* value is 5.5, degree of freedom is 198 and the p-value is smaller than 0.0001. We conclude that the improvement is statistically significant. Hence, KIRSF performs better than RSF in terms of prediction accuracy. The survival functions for each individual in the training set using RSF and KIRSF are shown in Figure 6 and Figure 7, respectively. The overall ensemble survival and Nelson-Aalen estimator are also shown in these two figures.

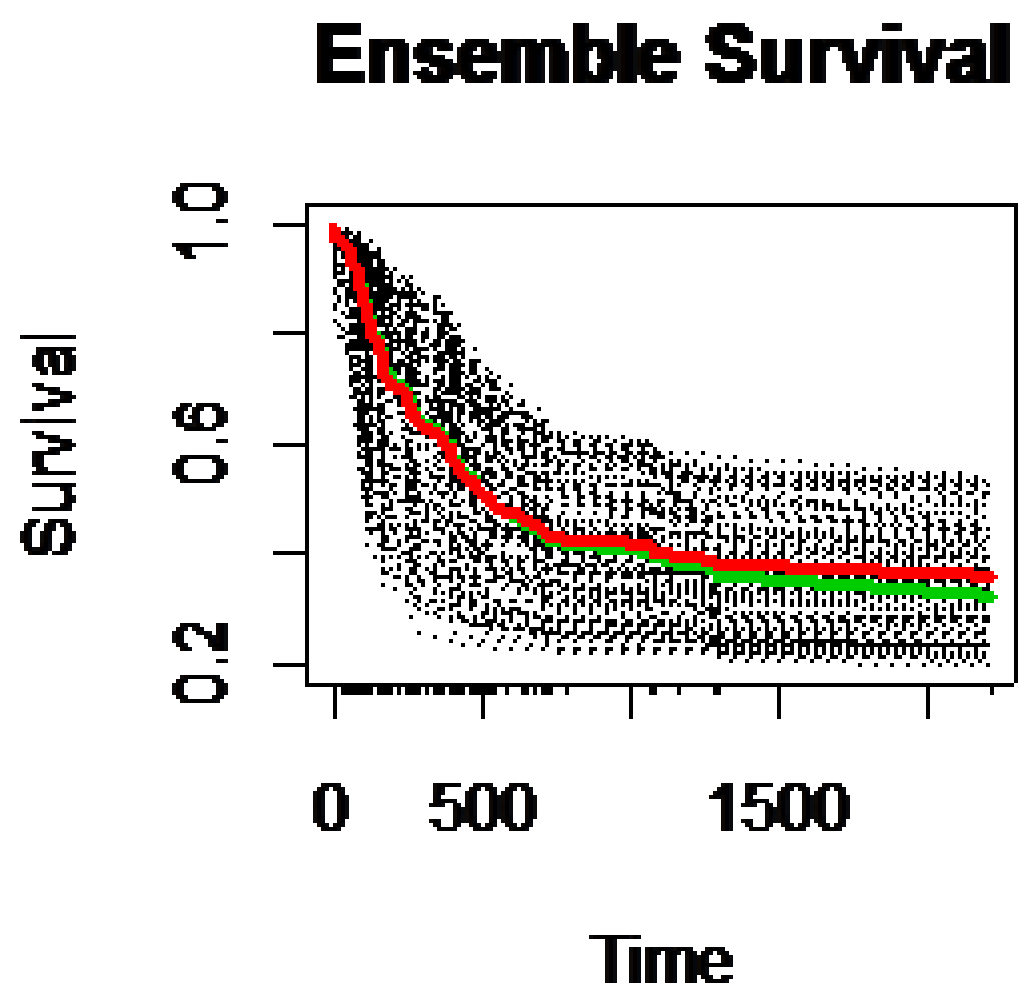

**Figure 6. Survival function: Red line is RSF overall ensemble survival . Green line is Nelson-Aalen estimator. Black lines are RSF survival functions for each individual in the training set.**

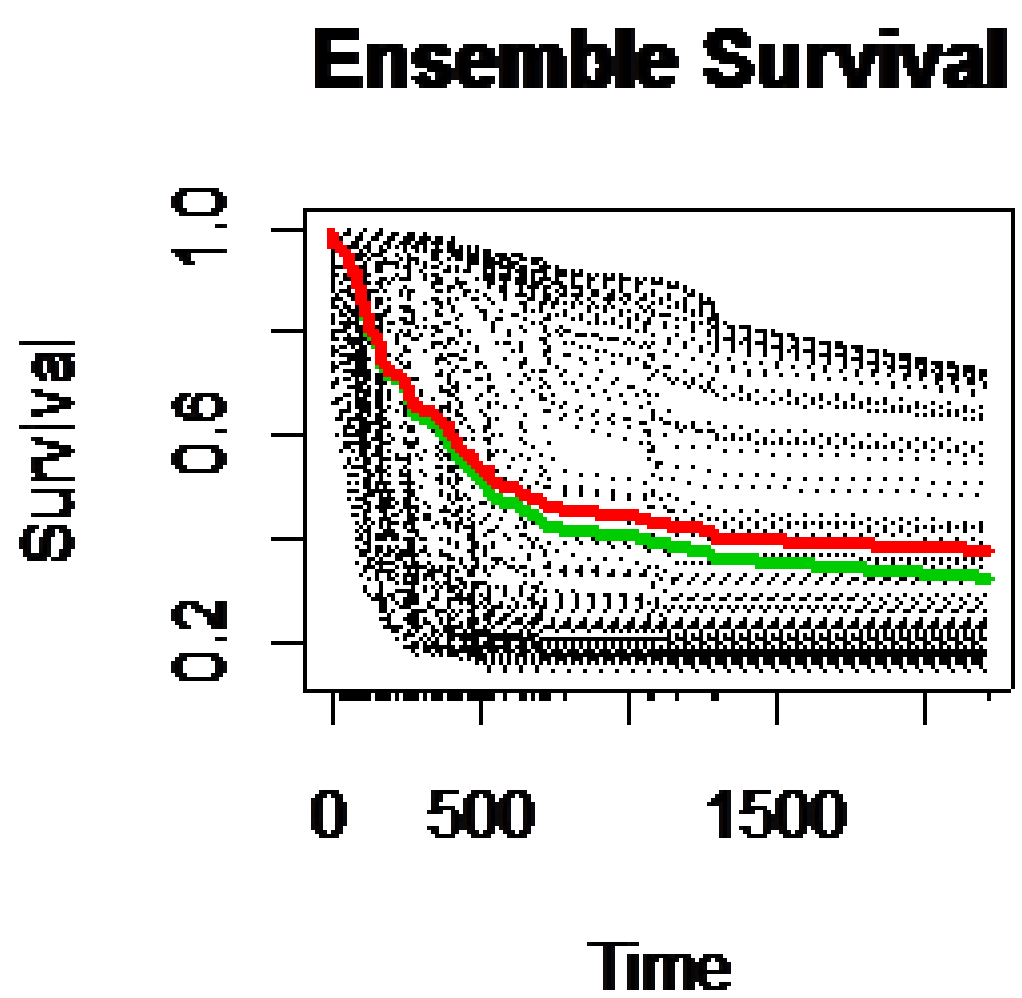

**Figure 7. Survival function: Red line is KIRSF overall ensemble survival. Green line is Nelson-Aalen estimator. Black lines are KIRSF survival functions for each individual in the training set.**

Figure 5 and Figure 6 show that both random survival forests overall ensemble survival and kernel induced random survival forests overall ensemble survival are very close to the Nelson-Aalen estimator. However, the kernel induced random survival function for each individual in the training set show that there may be two different survival groups in the population. One group has longer survival time than the other. Further investigation is needed to determine what characteristics separate the population into two different survival groups.

## 7. CONCLUSION

In this work we introduce how to incorporate flexible kernels into random survival forests while still keeping the random survival forests mechanism. We replace the original data by kernelized data and respect the random survival forests principle. Our method is applied to simulated ringnorm data set and real world data set: bone marrow transplant data. The experiments' results show that KIRSF reduce prediction error by 20 percent compared to RSF for the ringnorm data and over 30 percent for the BMT data. In addition, the sample standard error in KIRSF is smaller than those in Random Survival Forests. Hence, we suggest that if non-linear pattern exists in the data sets, using kernelized data instead of original data may decrease the prediction error greatly.

# 8. DISCUSSION

## 8.1 Dimension Reduction

In this paper, we consider all of the covariates when we apply kernel functions. However, when number of covariates is large, Computation efficiency is a concern. To eliminate noise covariates and keep a good computation load, it is necessary to conduct variable selections before we apply any kernels to survival data. Some variable selection techniques for survival data are listed as below: Lasso methods were more accurate than stepwise selection [5]. A Bayesian variable selection method for censored data [6] approximated the posterior distribution of the parameters for the proportional hazard model by the maximum partial likelihood estimator. Bayesian information criterion [7] was proposed by modifying penalty term in the criterion for predictor variable selection. A predictor variable selection method was proposed through non-concave penalized likelihood [8]. Hence, one of the variable selection techniques could be used before applying kernel functions to the data set. By reducing the number of covariates, we can not only improve computing efficiency, but also can take better advantage of kernel functions.

## 8.2 Another Future Work

We demonstrate how to incorporate kernel functions into Random Survival Forests. The implementation of our method to a wide range of data sets should be investigated. We only apply our method to Ringnorm and BMT data. Hence, it would be interesting to find out the patterns of data sets in which better prediction accuracy can be achieved by using different types of kernel functions.